# BUSINESS INTELLIGENCE FROM WEB USAGE MINING

Ajith Abraham

Department of Computer Science, Oklahoma State University,
700 N Greenwood Avenue, Tulsa,Oklahoma 74106-0700, USA,
*ajith.abraham@ieee.org*

**Abstract.** The rapid e-commerce growth has made both business community and customers face a new situation. Due to intense competition on one hand and the customer's option to choose from several alternatives business community has realized the necessity of intelligent marketing strategies and relationship management. Web usage mining attempts to discover useful knowledge from the secondary data obtained from the interactions of the users with the Web. Web usage mining has become very critical for effective Web site management, creating adaptive Web sites, business and support services, personalization, network traffic flow analysis and so on. In this paper, we present the important concepts of Web usage mining and its various practical applications. We further present a novel approach 'intelligent-miner' (*i-Miner*) to optimize the concurrent architecture of a fuzzy clustering algorithm (to discover web data clusters) and a fuzzy inference system to analyze the Web site visitor trends. A hybrid evolutionary fuzzy clustering algorithm is proposed in this paper to optimally segregate similar user interests. The clustered data is then used to analyze the trends using a Takagi-Sugeno fuzzy inference system learned using a combination of evolutionary algorithm and neural network learning. Proposed approach is compared with self-organizing maps (to discover patterns) and several function approximation techniques like neural networks, linear genetic programming and Takagi-Sugeno fuzzy inference system (to analyze the clusters). The results are graphically illustrated and the practical significance is discussed in detail. Empirical results clearly show that the proposed Web usage-mining framework is efficient.

## 1. Introduction

The WWW continues to grow at an amazing rate as an information gateway and as a medium for conducting business. Web mining is the extraction of interesting and useful knowledge and implicit information from atrifacts or activity related to the WWW [23][14]. Based on several reserch studies we can broadly classify Web mining into three domains: content, structure and usage mining [8][9]. The discussions in this chapter will be limited to Web usage mining. Web servers record and accumulate data about user interactions whenever requests for resources are received. Analyzing the Web access logs can help understand the user behaviour and the web structure. From the business and applications point of view, knowledge obtained from the Web usage patterns could be directly applied to efficiently manage activities related to e-business, e-services, e-education and so on [10][11]. Accurate Web usage information could help to attract new customers, retain current customers, improve cross marketing/sales, effectiveness of promotional campaigns, tracking leaving customers and find the most effective logical structure for their Web space [19]. User profiles could be built by combining users' navigation paths with other data features, such as page viewing time, hyperlink structure, and page content [17]. What makes the discovered knowledge interesting had been addressed by several works. Results previously known are very often considered as not interesting. So the key concept to make the discovered knowledge interesting will be its novelty or unexpectedness appearance [4][5][13].

When ever a visitor access the server it leaves the IP, authenticated user ID, time/date, request mode, status, bytes, referrer, agent and so on. The available data fields are specified by the HTTP protocol. There are several commercial softwares that could provide Web usage ststistics. These stats could be useful for Web administrators to get a sense of the actual load on the server. For small web servers, the usage statistics provided by conventional Web site trackers may be adequate to analyze the usage pattern and trends. However as the size and complexity of the data increases, the statistics provided by existing Web log file analysis tools may prove inadequate and more intelligent mining techniques will be necessary [20].

In the case of Web mining, data could be collected at the server level, client level, proxy level or some consolidated data. These data could differ in terms of content and the way it is collected etc. The usage data collected at different sources represent the navigation patterns of different segments of the overall Web traffic, ranging from single user, single site browsing behaviour to multi-user, multi-site access patterns. Web server log does not accurately contain sufficient information for infering the behaviour at the client side as they relate to the pages served by the Web server. Pre-procesed and cleaned data could be used for pattern discovery, pattern analysis, Web usage ststistics and generating association/ sequential rules. Much work has been performed on extracting various pattern information from Web logs and the application of the discovered knowledge range from improving the design and structure of a Web site to enabling business organizations to function more effeciently [22][24][27][28][29][30][31][33].

Jespersen et al [20] proposed an hybrid approach for analyzing the visitor click sequences. A combination of hypertext probabilistic grammar and click fact table approach is used to mine Web logs which could be also used for general sequence mining tasks. Mobasher et al [25] proposed the Web personalization system which consists of offline tasks related to the mining of usage data and online process of automatic Web page customization based on the knowledge discovered. LOGSOM proposed by Smith et al [32], utilizes self-organizing map to organize web pages into a two-dimensional map based solely on the users' navigation behavior, rather than the content of the web pages. LumberJack proposed by Chi et al [12] builds up user profiles by combining both user session clustering and traditional statistical traffic analysis using K-means algorithm. Joshi et al [21] used relational online analytical processing approach for creating a Web log warehouse using access logs and mined logs (association rules and clusters). A comprehensive overview of Web usage mining research is found in [14][34].

To demonstrate the effeciency of the proposed frameworks, Web access log data at the Monash University's Web site [26] were used for experimentations. The University's central web server receives over 7 million hits in a week and therefore it is a real challenge to find and extract hidden usage pattern information. The average daily and hourly patterns even though tend to follow a similar trend (as evident from the figures) the differences tend to increase during high traffic days (Monday – Friday) and during the peak hours (11:00-17:00 Hrs). Due to the enormous traffic volume and chaotic access behavior, the prediction of the user access patterns becomes more difficult and complex.

Self organizing maps and fuzzy c-means algorithm could be used to seggregate the user access records and computational intelligence paradigms to analyze the user access trends. Experimentation results [3][36] have clearly shown the importance of the clustering algorithm to analyze the user access trends.

In the subsequent section, we present some theoretical concepts of clustering algorithms and various computational intelligence paradigms. Experimentation results are provided in Section 3 and some conclusions are provided towards the end.

## 2. Mining Framework Using Hybrid Computational Intelligence Paradigms (CI)

### 2.1 Clustrering Algorithms

*Fuzzy Clustering Algorithm*

One of the widely used clustering methods is the fuzzy c-means (FCM) algorithm developed by Bezdek [7]. FCM partitions a collection of $n$ vectors $x_i$, $i= 1,2...,n$ into $c$ fuzzy groups and finds a cluster center in each group such that a cost function of dissimilarity measure is minimized. To accommodate the introduction of fuzzy partitioning, the membership matrix $U$ is allowed to have elements with values between 0 and 1. The FCM objective function takes the form

$$J(U,c_1,...c_c) = \sum_{i=1}^{c} J_i = \sum_{i=1}^{c} \sum_{j=1}^{n} u_{ij}^m d_{ij}^2 \qquad (1)$$

Where $u_{ij}$, is a numerical value between [0,1]; $c_i$ is the cluster center of fuzzy group $i$; $d_{ij} = \|c_i - x_j\|$ is the Euclidian distance between $i^{th}$ cluster center and $j^{th}$ data point; and $m$ is called the exponential weight which influences the degree of fuzziness of the membership (partition) matrix.

*Self Organizing Map (SOM)*

The SOM is an algorithm used to visualize and interpret large high-dimensional data sets. The map consists of a regular grid of processing units, "neurons". A model of some multidimensional observation, eventually a vector consisting of features, is associated with each unit. The map attempts to represent all the available observations with optimal accuracy using a restricted set of models. At the same time the models become ordered on the grid so that similar models are close to each other and dissimilar models far from each other.

Fitting of the model vectors is usually carried out by a sequential regression process, where $t = 1,2,...$ is the step index: For each sample $x(t)$, first the winner index $c$ (best match) is identified by the condition

$$\forall_i, \|x(t) - m_c(t)\| \leq \|x(t) - m_i(t)\| \qquad (2)$$

After that, all model vectors or a subset of them that belong to nodes centered around node $c = c(x)$ are updated as

$$m_i(t+1) = m_i(t) + h_{c(x)i}(x(t) - m_i(t)) \qquad (3)$$

Here $h_{c(x)i}$ is the neighborhood function, a decreasing function of the distance between the $i^{th}$ and $c^{th}$ nodes on the map grid. This regression is usually reiterated over the available samples.

## 2.2 Computational Intelligence (CI)

CI substitutes intensive computation for insight into how complicated systems work. Artificial neural networks, fuzzy inference systems, probabilistic computing, evolutionary computation etc were all shunned by classical system and control theorists. CI provides an excellent framework unifying them and even by incorporating other revolutionary methods.

*Artificial Neural Network (ANN)*

ANNs were designed to mimic the characteristics of the biological neurons in the human brain and nervous system. Learning typically occurs by example through training, where the training algorithm iteratively adjusts the connection weights (synapses). Backpropagation (BP) is one of the most famous training algorithms for multilayer perceptrons. BP is a gradient descent technique to minimize the error $E$ for a particular training pattern. For adjusting the weight ($w_{ij}$) from the $i^{th}$ input unit to the $j^{th}$ output, in the batched mode variant the descent is based on the gradient $\nabla E$ ($\frac{\delta E}{\delta w_{ij}}$) for the total training set

$$\Delta w_{ij}(n) = -\varepsilon * \frac{\delta E}{\delta w_{ij}} + \alpha * \Delta w_{ij}(n-1) \qquad (4)$$

The gradient gives the direction of error $E$. The parameters $\varepsilon$ and $\alpha$ are the learning rate and momentum respectively.

*Linear Genetic Programming (LGP)*

Linear genetic programming is a variant of the GP technique that acts on linear genomes [6]. Its main characteristics in comparison to tree-based GP lies in that the evolvable units are not the expressions of a functional programming language (like LISP), but the programs of an imperative language (like c/c ++). An alternate approach is to evolve a computer program at the machine code level, using lower level representations for the individuals. This can tremendously hasten up the evolution process as, no matter how an individual is initially represented, finally it always has to be represented as a piece of machine code, as fitness evaluation requires physical execution of the individuals.

The basic unit of evolution here is a native machine code instruction that runs on the floating-point processor unit (FPU). Since different instructions may have different sizes, here instructions are clubbed up together to form instruction blocks of 32 bits each. The instruction blocks hold one or more native machine code instructions, depending on the sizes of the instructions. A crossover point can occur only between instructions and is prohibited from occurring within an instruction. However the mutation operation does not have any such restriction.

*Fuzzy Inference Systems (FIS)*

Fuzzy logic provides a framework to model uncertainty, human way of thinking, reasoning and the perception process. Fuzzy *if-then* rules and fuzzy reasoning are the backbone of fuzzy inference systems, which are the most important modelling tools based on fuzzy set theory. We made use of the Takagi Sugeno fuzzy inference scheme in which the conclusion of a fuzzy rule is constituted by a weighted linear combination of the crisp inputs rather than a fuzzy set [35]. In our simulation, we used Adaptive Network Based Fuzzy Inference System (ANFIS) [18], which implements a Takagi Sugeno fuzzy inference system.

*Optimization of Fuzzy Clustering Algorithm*

*Optimization of* Usually a number of cluster centers are randomly initialized and the FCM algorithm provides an iterative approach to approximate the minimum of the objective function starting from a given position and leads to any of its local minima [7]. No guarantee ensures that FCM converges to an optimum solution (can be trapped by local extrema in the process of optimizing the clustering criterion). The performance is very sensitive to initialization of the cluster centers. An evolutionary algorithm is used to decide the optimal number of clusters and their cluster centers. The algorithm is initialized by constraining the initial values to be within the space defined by the vectors to be clustered. A very similar approach is given in [16].

*Optimization of Fuzzy Inference System*

We used the EvoNF framework [2], which is an integrated computational framework to optimize fuzzy inference system using neural network learning and evolutionary computation. Solving multi-objective scientific and engineering problems is, generally, a very difficult goal. In these particular optimization problems, the objectives often conflict across a high-dimension problem space and may also require extensive computational resources. The hierarchical evolutionary search framework could adapt the membership functions (shape and quantity), rule base (architecture), fuzzy inference mechanism (T-norm and T-conorm operators) and the learning parameters of neural network learning algorithm [1]. In addition to the evolutionary learning (global search) neural network learning could be considered as a local search technique to optimize the parameters of the rule antecedent/consequent parameters and the parameterized fuzzy operators. The hierarchical search could be formulated as follows:

For every fuzzy inference system, there exist a global search of neural network learning algorithm parameters, parameters of the fuzzy operators, *if-then* rules and membership functions in an environment decided by the problem. The evolution of the fuzzy inference system will evolve at the slowest time scale while the evolution of the quantity and type of membership functions will evolve at the fastest rate. The function of the other layers could be derived similarly. Hierarchy of the different adaptation layers (procedures) will rely on the prior knowledge (this will also help to reduce the search space). For example, if we know certain fuzzy operators will work well for a problem then it is better to implement the search of fuzzy operators at a higher level. For fine-tuning the fuzzy inference system all the node functions are to be parameterized. For example, the Schweizer and Sklar's T-norm operator can be expressed as:

$$T(a,b,p) = \left[ \max\left\{0, (a^{-p} + b^{-p} - 1)\right\} \right]^{-\frac{1}{p}} \tag{5}$$

It is observed that

$$\begin{aligned} \lim_{p \to 0} T(a,b,p) &= ab \\ \lim_{p \to \infty} T(a,b,p) &= \min\{a,b\} \end{aligned} \tag{6}$$

which correspond to two of the most frequently used T-norms in combining the membership values on the premise part of a fuzzy *if-then* rule.

### 2.3 Mining Framework Using Integrated Systems (*i-Miner*)

The hybrid framework optimizes a fuzzy clustering algorithm using an evolutionary algorithm and a Takagi-Sugeno fuzzy inference system using a combination of evolutionary algorithm and neural network learning. The raw data from the log files are cleaned and pre-processed and a fuzzy C means algorithm is used to identify the number of clusters [3].

The developed clusters of data are fed to a Takagi-Sugeno fuzzy inference system to analyze the trend patterns. The *if-then* rule structures are learned using an iterative learning procedure [15] by an evolutionary algorithm and the rule parameters are fine-tuned using a backpropagation algorithm.

The hierarchical distribution of the *i-Miner* is depicted in Figure 2. The arrow direction depicts the speed of the evolutionary search. The optimization of clustering algorithm progresses at a faster time scale in an environment decided by the inference method and the problem environment.

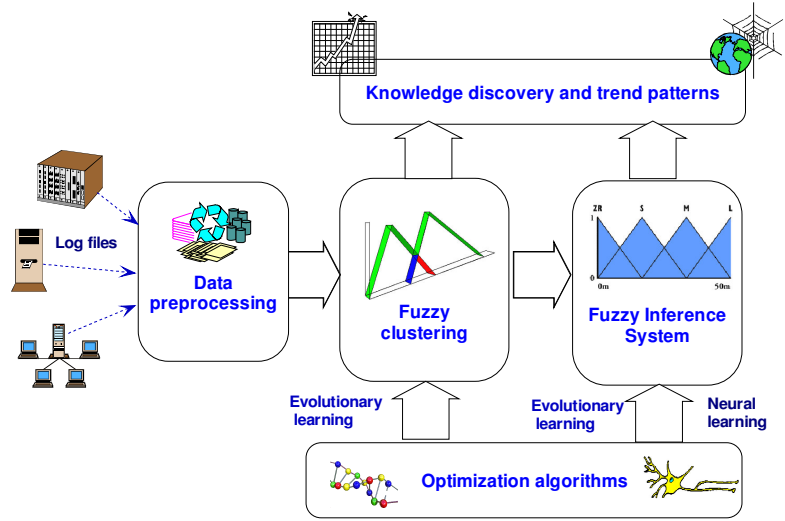

**Figure 1.** *i-Miner* framework

*Chromosome Modeling and Representation*

Hierarchical evolutionary search process has to be represented in a chromosome for successful modeling of the *i-Miner* framework. A typical chromosome of the *i-Miner* would appear as shown in Figure 3 and the detailed modeling process is as follows.

**Layer 1.** The optimal number of clusters and initial cluster centers is represented this layer.

**Layer 2.** This layer is responsible for the optimization of the rule base. This includes deciding the total number of rules, representation of the antecedent and consequent parts. The number of rules grows rapidly with an increasing number of variables and fuzzy sets. We used the grid-partitioning algorithm to generate the initial set of rules. An iterative learning method is then adopted to optimize the rules [15]. The existing rules are mutated and new rules are introduced. The fitness of a rule is given by its contribution (strength) to the actual output. To represent a single rule a position dependent code with as many elements as the number of variables of the system is used. Each element is a binary string with a bit per fuzzy set in the fuzzy partition of the variable, meaning the absence or presence of the corresponding linguistic label in the rule.

**Layer 3.** This layer is responsible for the selection of optimal learning parameters. Performance of the gradient descent algorithm directly depends on the learning rate according to the error surface. The optimal learning parameters decided by this layer will be used to tune the parameterized rule antecedents/consequents and the fuzzy operators.

The rule antecedent/consequent parameters and the fuzzy operators are fine tuned using a gradient descent algorithm to minimize the output error

$$E = \sum_{k=1}^{N}(d_k - x_k)^2 \qquad (7)$$

where $d_k$ is the $k^{th}$ component of the $r^{th}$ desired output vector and $x_k$ *is the* $k^{th}$ component of the actual output vector by presenting the $r^{th}$ input vector to the network. All the gradients of the parameters to be optimized, namely the consequent parameters $\frac{\partial E}{\partial P_n}$ for all rules $R_n$ and the premise parameters $\frac{\partial E}{\partial \sigma_i}$ and $\frac{\partial E}{\partial c_i}$ for all fuzzy sets $F_i$ ($\sigma$ and c represents the MF width and center of a Gaussian MF).

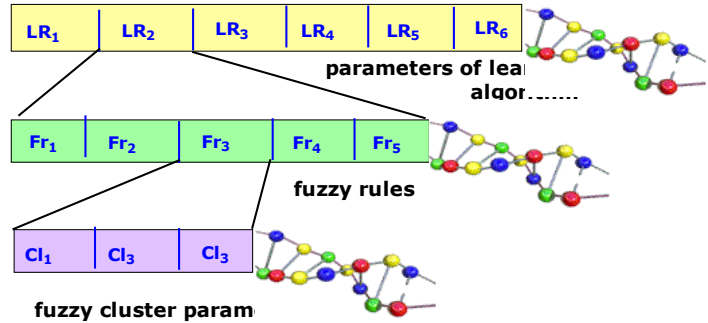

**Figure 2.** Chromosome structure of the *i-Miner*

Once the three layers are represented in a chromosome *C*, and then the learning procedure could be initiated as follows:
  a. Generate an initial population of N numbers of C chromosomes. Evaluate the fitness of each chromosome depending on the output error.
  b. Depending on the fitness and using suitable selection methods reproduce a number of children for each individual in the current generation.
  c. Apply genetic operators to each child individual generated above and obtain the next generation.
  d. Check whether the current model has achieved the required error rate or the specified number of generations has been reached. Go to Step b.
  e. End

## 3. Experimentation Setup-Training and Performance Evaluation

In this research, we used the statistical/ text data generated by the log file analyzer from 01 January 2002 to 07 July 2002. Selecting useful data is an important task in the data pre-processing block. After some preliminary analysis, we selected the statistical data comprising of domain byte requests, hourly page requests and daily page requests as focus of the cluster models for finding Web users' usage patterns. It is also important to remove irrelevant and noisy data in order to build a precise model. We also included an additional input '*index number*' to distinguish the time sequence of the data. The most recently accessed data were indexed higher while the least recently accessed data were placed at the bottom. Besides the inputs '*volume of requests*' and '*volume of pages (bytes)*' and '*index number*', we also used the '*cluster information*' provided by the clustering algorithm as an additional input variable. The data was re-indexed based on the cluster information. Our task is to predict (few time steps ahead) the Web traffic volume on a hourly and daily basis. We used the data from 17 February 2002 to 30 June 2002 for training and the data from 01 July 2002 to 06 July 2002 for testing and validation purposes.

**Table 1.** Parameter settings of *i-Miner*

| | |
|---|---|
| Population size | 30 |
| Maximum no of generations | 35 |
| Fuzzy inference system | Takagi Sugeno |
| Rule antecedent membership functions | 3 membership functions per input variable (parameterized Gaussian) |
| Rule consequent parameters | linear parameters |
| Gradient descent learning | 10 epochs |
| Ranked based selection | 0.50 |
| Elitism | 5 % |
| Starting mutation rate | 0.50 |

The initial populations were randomly created based on the parameters shown in Table 1. We used a special mutation operator, which decreases the mutation rate as the algorithm greedily proceeds in the search space [15]. If the allelic value $x_i$ of the $i$-th gene ranges over the domain $a_i$ and $b_i$ the mutated gene $x_i'$ is drawn randomly uniformly from the interval $[a_i, b_i]$.

$$x_i' = \begin{cases} x_i + \Delta(t, b_i - x_i), & \text{if } \omega = 0 \\ x_i + \Delta(t, x_i - a_i), & \text{if } \omega = 1 \end{cases} \quad (8)$$

where $\omega$ represents an unbiased coin flip $p(\omega=0) = p(\omega=1) = 0.5$, and

$$\Delta(t, x) = x \left( 1 - \gamma^{\left(1 - \frac{t}{t_{max}}\right)^b} \right) \quad (9)$$

defines the mutation step, where $\gamma$ is the random number from the interval $[0,1]$ and $t$ is the current generation and $t_{max}$ is the maximum number of generations. The function $\Delta$ computes a value in the range $[0,x]$ such that the probability of returning a number close to zero increases as the algorithm proceeds with the search. The parameter $b$ determines the impact of time on the probability distribution $\Delta$ over $[0,x]$. Large values of $b$ decrease the likelihood of large mutations in a small number of generations. The parameters mentioned in Table 1 were decided after a few trial and error approaches. Experiments were repeated 3 times and the average performance measures are reported. Figures 3 and 4 illustrates the meta-learning approach combining evolutionary learning and gradient descent technique during the 35 generations.

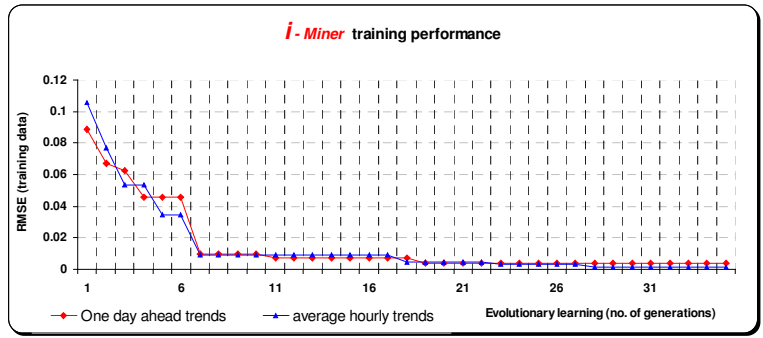

**Figure 3.** Meta-learning performance (training) of *i-Miner*

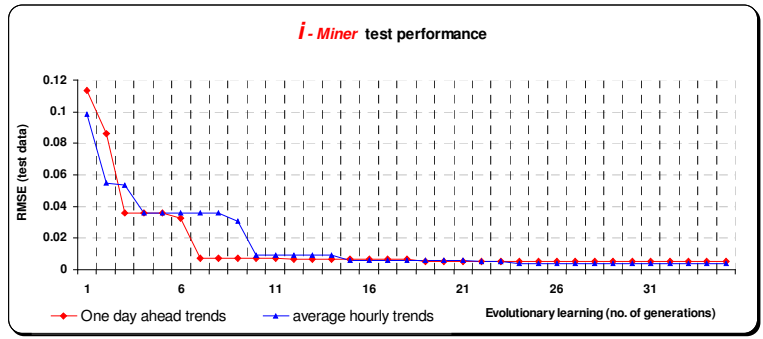

**Figure 4.** Meta-learning performance (testing) of *i-Miner*

Table 2 summarizes the performance of the developed *i-Miner* for training and test data. Performance is compared with the previous results [36][27] wherein the trends were analyzed using a Takagi-Sugeno Fuzzy

Inference System (ANFIS), Artificial Neural Network (ANN) and Linear Genetic Programming (LGP). The Correlation Coefficient (CC) for the test data set is also given in Table 2. The 35 generations of meta-learning approach created 62 *if-then* Takagi-Sugeno type fuzzy rules (daily traffic trends) and 64 rules (hourly traffic trends) compared to the 81 rules reported in [36].

Figures 5 and 6 illustrate the actual and predicted trends for the test data set. A trend line is also plotted using a least squares fit (6$^{th}$ order polynomial).

FCM approach created 7 data clusters for hourly traffic according to the input features compared to 9 data clusters for the daily requests. The previous study using Self-organizing Map (SOM) created 7 data clusters (daily traffic volume) and 4 data clusters (hourly traffic volume) respectively. As evident, FCM approach resulted in the formation of additional data clusters. Several meaningful information could be obtained from the clustered data. Depending on the volume of requests and transfer of bytes, data clusters were formulated. Clusters based on hourly data show the visitor information at certain hour of the day.

**Table 2.** Performance of the different paradigms

| Method | Period | | | | | |
|---|---|---|---|---|---|---|
| | Daily (1 day ahead) | | | Hourly (1 hour ahead) | | |
| | RMSE | | CC | RMSE | | CC |
| | Train | Test | | Train | Test | |
| i-Miner | 0.0044 | 0.0053 | 0.9967 | 0.0012 | 0.0041 | 0.9981 |
| TKFIS | 0.0176 | 0.0402 | 0.9953 | 0.0433 | 0.0433 | 0.9841 |
| ANN | 0.0345 | 0.0481 | 0.9292 | 0.0546 | 0.0639 | 0.9493 |
| LGP | 0.0543 | 0.0749 | 0.9315 | 0.0654 | 0.0516 | 0.9446 |

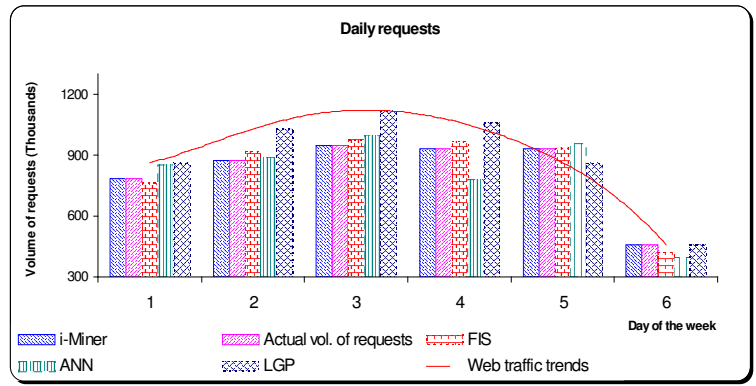

**Figure 5.** Test results of the daily trends for 6 days

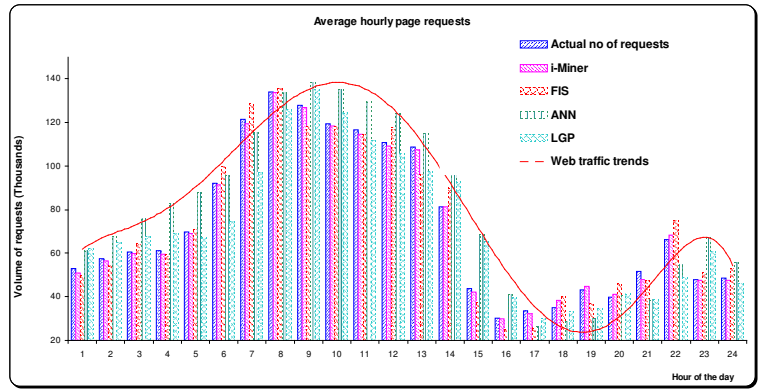

**Figure 6.** Test results of the average hourly trends for 6 days

## 4. Conclusions

Recently Web usage mining has been gaining a lot of attention because of its potential commercial benefits. The proposed *i-Miner* framework seems to work very well for the problem considered. The empirical results also reveal the importance of using soft computing paradigms for mining useful information. In this chapter, our focus was to develop accurate trend prediction models to analyze the hourly and daily web traffic volume. Several useful information could be discovered from the clustered data. FCM clustering resulted in more clusters compared to SOM approach. Perhaps more clusters were required to improve the accuracy of the trend analysis. The main advantage of SOMs comes from the easy visualization and interpretation of clusters formed by the map. The knowledge discovered from the developed FCM clusters and SOM could be a good comparison study and is left as a future research topic.

As illustrated in Table 2, *i-Miner* framework gave the overall best results with the lowest RMSE on test error and the highest correlation coefficient. It is interesting to note that the three considered soft computing paradigms could easily pickup the daily and hourly Web-access trend patterns. When compared to LGP, the developed neural network performed better (in terms of RMSE) for daily trends but for hourly trends LGP gave better results. An important disadvantage of *i-Miner* is the computational complexity of the algorithm. When optimal performance is required (in terms of accuracy and smaller structure) such algorithms might prove to be useful as evident from the empirical results.

So far most analysis of Web data have involved basic traffic reports that do not provide much pattern and trend analysis. By linking the Web logs with cookies and forms, it is further possible to analyze the visitor behavior and profiles which could help an e-commerce site to address several business questions. Our future research will be oriented in this direction by incorporating more data mining paradigms to improve knowledge discovery and association rules from the clustered data.